\documentclass[letterpaper]{article}
\usepackage{aaai20}
\usepackage{times}
\usepackage{helvet}
\usepackage{courier}

\usepackage[T1]{fontenc}
\usepackage[utf8]{inputenc}

\usepackage{latexsym}
\usepackage{float}
\restylefloat{table}
\usepackage{url}
\usepackage{graphicx}

\usepackage{makecell}
\usepackage{todonotes}
 
\begin{document}
%

\title{Linguistic Fingerprints of Internet Censorship: the Case of  Sina Weibo} 
\author{Kei Yin Ng, Anna Feldman, Jing Peng\\
Montclair State University\\
Montclair, New Jersey, USA\\
}
\maketitle
\begin{abstract}
\begin{quote}
This paper studies how the linguistic components of blogposts collected from Sina Weibo, a Chinese microblogging platform, might affect the blogposts' likelihood of being censored. 
Our results go along with King et al.~\shortcite{king-etal:2013}'s Collective Action Potential (CAP) theory, which states that a blogpost's potential of causing riot or assembly in real life is the key determinant of it getting censored. Although there is not a definitive measure of this construct,  the linguistic features that we identify as discriminatory go along with the CAP theory. We build a classifier that significantly outperforms  non-expert humans in predicting whether a blogpost will be censored.  The crowdsourcing results suggest that while humans tend to see censored blogposts as more controversial and more likely to trigger action in real life than the uncensored counterparts, they in general cannot make a better guess than our model when it comes to `reading the mind' of the censors in deciding whether a blogpost should be censored.  We do not claim that censorship is only determined by the linguistic features.  There are many other factors contributing to censorship decisions. The focus of the present paper is on the linguistic form of blogposts. Our work suggests that it is possible to use linguistic properties of social media posts to automatically predict if they are going to be censored.
\end{quote}
\end{abstract}

\section{Introduction}

In 2019, Freedom in the World\footnote{\url{https://freedomhouse.org/report/freedom-world/freedom-world-2019/democracy-in-retreat}}, a yearly survey produced by Freedom House\footnote{\url{https://freedomhouse.org/}}  that measures the degree of civil liberties and political rights in every nation, recorded the 13th consecutive year of decline in global freedom. 
This decline spans across long-standing democracies such as USA as well as authoritarian regimes such as China and Russia. ``Democracy is in retreat. The offensive against freedom of expression is being supercharged by a new and more effective form of digital authoritarianism." 
According to the report, China is now exporting its model of comprehensive internet censorship and surveillance around the world, offering trainings, seminars, and even study trips as well as advanced equipment.  


In this paper, we deal with a particular type of censorship -- when a post gets removed from a social media platform semi-automatically based on its content. We are interested in exploring whether there are systematic linguistic differences between posts that get removed by censors from Sina Weibo,  a Chinese microblogging platform, and the posts that remain on the website. Sina Weibo was launched in 2009 and became the most popular social media platform in China. Sina Weibo has over 431 million monthly active users\footnote{\url{https://www.investors.com/news/technology/weibo-reports-first-quarter-earnings/}}.

In cooperation with the ruling regime, Weibo sets strict control over the content published under its service \cite{pen-2018}. According to Zhu et al. \shortcite{zhu-etal:2013}, Weibo uses a variety of strategies to target censorable posts, ranging from keyword list filtering to individual user monitoring. Among all posts that are eventually censored, nearly 30\% of them are censored within 5--30 minutes, and nearly 90\% within 24 hours \cite{zhu-etal:2013}. We hypothesize that the former are done automatically, while the latter are removed by human censors.  

 Research shows that some of the censorship decisions are not necessarily driven by the criticism of the state \cite{king-etal:2013}, the presence of controversial topics \cite{ng2018detecting,kei-nlp4if:2018}, or posts that describe negative events \cite{ng-etal-2019-neural}. Rather, censorship is triggered by other factors, such as for example, the collective action potential \cite{king-etal:2013}, i.e., censors target posts that stimulate collective action, such as 
 riots and protests.
 
The goal of this paper is to compare censored and uncensored posts that contain the same sensitive keywords and topics. Using the linguistic features extracted, a neural network model is built to explore whether censorship decision can be deduced from the linguistic characteristics of the posts.

The contributions of this paper are: 
1. We decipher a way to determine whether a blogpost on Weibo has been deleted by the author or censored by Weibo. 
2. We develop a corpus of censored and uncensored Weibo blogposts that contain sensitive keyword(s).
3. We build a neural network classifier that predicts censorship significantly better than non-expert humans.
4. We find a set of linguistics features that contributes to the censorship prediction problem. 
5. We indirectly test the construct of Collective Action Potential (CAP) proposed by King et al. \shortcite{king-etal:2013} through crowdsourcing experiments and find that the existence of CAP is more prevalent in censored blogposts than uncensored blogposts as judged by human annotators.  


\section{Previous Work}

There have been significant efforts to develop strategies to detect and evade censorship. Most work, however, focuses on exploiting technological limitations with existing routing protocols \cite{leberknight-etal:2012a,katti-etal:2005,levin-etal:2015,mcpherson-etal:2016,weinberg-etal:2012}. Research that pays more attention to linguistic properties of online censorship in the context of censorship evasion include, for example, Safaka et al.~\shortcite{safaka-etal:2016} who apply linguistic steganography to circumvent censorship. Lee~\shortcite{lee:2016} uses parodic satire to bypass censorship in China and claims that this stylistic device delays and often evades censorship. Hiruncharoenvate et al.~\shortcite{hirun-etal:2015} show that the use of homophones of censored keywords on Sina Weibo could help extend the time a Weibo post could remain available online. All these methods rely on a significant amount of human effort to interpret and annotate texts to evaluate the likelihood of censorship, which might not be practical to carry out for common Internet users in real life.  There has also been research that uses linguistic and content clues to detect censorship. Knockel et al.~\shortcite{knockel-etal:2015} and Zhu et al.~\shortcite{zhu-etal:2013} propose detection mechanisms to categorize censored content and automatically learn keywords that get censored. King et al.~\shortcite{king-etal:2013} in turn study the relationship between political criticism and chance of censorship. They come to the conclusion that posts that have a Collective Action Potential get deleted by the censors even if they support the state. Bamman et al.~\shortcite{bamman-etal:2012} uncover a set of politically sensitive keywords and find that the presence of some of them in a Weibo blogpost contribute to a  higher chance of the post being censored. Ng et al.~\shortcite{kei-nlp4if:2018} also target a set of topics that have been suggested to be sensitive, but unlike Bamman et al.~\shortcite{bamman-etal:2012}, they cover areas not limited to politics. 
Ng et al.~\shortcite{kei-nlp4if:2018} investigate how the textual content as a whole might be relevant to censorship decisions when both the censored and uncensored blogposts include the same sensitive keyword(s). 

Our work is related to Ng et al.~\shortcite{kei-nlp4if:2018} and Ng et al.~\shortcite{ng-etal-2019-neural}; however, we introduce a larger and more diverse dataset of censored posts; we experiment with a wider range of features and in fact show that not all the features reported in Ng et al. guarantee the best performance. We built a classifier that significantly outperforms Ng et al. We conduct a crowdsourcing experiment testing human judgments of controversy and censorship  as well as indirectly testing the construct of collective action potential proposed by King et al.

\section{Tracking Censorship} \label{sec:collection}
Tracking censorship topics on Weibo is a challenging task due to the transient nature of censored posts and the scarcity of censored data from well-known sources such as FreeWeibo\footnote{\url{https://freeweibo.com/}} and WeiboScope\footnote{\url{http://weiboscope.jmsc.hku.hk/}}. The most straightforward way to collect data from a social media platform is to make use of its API. However, Weibo imposes various restrictions on the use of its API such as restricted access to certain endpoints and restricted number of posts returned per request. Above all, the Weibo API does not provide any endpoint that allows easy and efficient collection of the target data (posts that contain sensitive keywords). Therefore, an alternative method is needed to track censorship for our purpose. 
\section{Datasets}

\subsection{Using Zhu et al. (2003)'s Corpus}
Zhu et al. \shortcite{zhu-etal:2013} collected over 2 million posts published by a set of around 3,500 sensitive users during a 2-month period in 2012. We extract around 20 thousand text-only posts using 64 keywords across 26 topics, which partially overlap with those included in the New Corpus (see below and in  Table \ref{tab:jed-table}).  We filter all duplicates. Among the extracted posts, 930 (4.63\%) are censored by Weibo as verified by Zhu et al. \shortcite{zhu-etal:2013} The extracted data from Zhu et al.\shortcite{zhu-etal:2013}'s are also used in building classification models. 

While it is possible to study the linguistic features in Zhu et al’s dataset without collecting new data, we created another corpus that targets `normal' users (Zhu et al. target `sensitive' users) and a different time period so that the results are not specific to a particular group of users and time. 

\subsection{New Corpus}
\subsubsection{Web Scraping}
We develop a web scraper that continuously collects and tracks data that contain sensitive keywords on the front-end. The scraper's target interface\footnote{e.g. searching "NLP"  http://s.weibo.com/weibo/NLP} displays 20 to 24 posts that contain a certain search key term(s), resembling a search engine's result page. We call this interface the Topic Timeline since the posts all contain the same keyword(s) and are displayed in reverse chronological order. The Weibo API does not provide any endpoint that returns the same set of data appeared on the Topic Timeline. Through a series of trial-and-errors to avoid CAPTCHAs that interrupt the data collection process, we found an optimal scraping frequency of querying the Topic Timeline every 5 to 10 minutes using 17 search terms (see Appendix\footnote{\url{https://msuweb.montclair.edu/~feldmana/publications/aaai20_appendix.pdf}}) across 10 topics (see Table \ref{tab:scraped-table}) for a period of 4 months (August 29, 2018 to December 29, 2018). In each query, all relevant posts and their meta-data are saved to our database. We save posts that contain texts only (i.e. posts that do not contain images, hyperlinks, re-blogged content etc.) and filter out duplicates.

\subsubsection{Decoding Censorship} \label{sec:decode}
\begin{figure*}[t]
 \begin{center}  
 \includegraphics[scale=0.4]{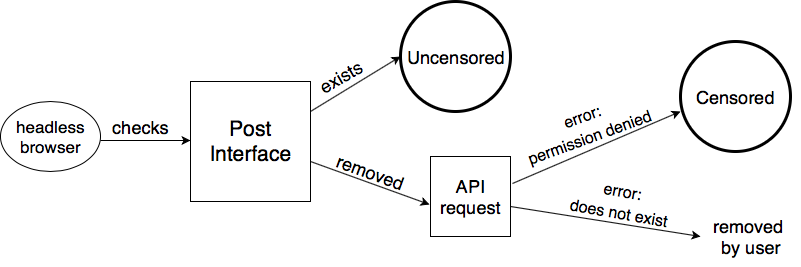}
 \caption{Logical flow to determine censorship status \cite{kei-nlp4if:2018}}\label{fig:flow}
\end{center} 
\end{figure*}

According to Zhu et al. \shortcite{zhu-etal:2013}, the unique ID of a Weibo post is the key to distinguish whether a post has been censored by Weibo or has been instead removed by the authors themselves.  
If a post has been censored by Weibo, querying its unique ID through the API returns an error message of ``permission denied" (system-deleted), whereas a user-removed post returns an error message of ``the post does not exist" (user-deleted). However, since the Topic Timeline (the data source of our web scraper) can be accessed only on the front-end (i.e. there is no API endpoint associated with it), we rely on both the front-end and the API to identify system- and user-deleted posts. It is not possible to distinguish the two types of deletion by directly querying the unique ID of all scraped posts because, through empirical experimentation, uncensored posts and censored (system-deleted) posts both return the same error message -- ``permission denied"). Therefore, we need to first check if a post still exists on the front-end, and then send an API request using the unique ID of the post that no longer exists to determine whether it has been deleted by the system or the user. 
The steps to identify censorship status of each post are illustrated in Figure \ref{fig:flow}. First, we check whether a scraped post is still available through visiting the user interface of each post. This is carried out automatically in a headless browser 2 days after a post is published. If a post has been removed (either by system or by user), the headless browser is redirected to an interface that says ``the page doesn't exist"; otherwise, the browser brings us to the original interface that displays the post content. Next, after 14 days, we use the same methods in step 1 to check the posts' status again. This step allows our dataset to include posts that have been removed at a later stage. Finally, we send a follow-up API query using the unique ID of posts that no longer exist on the browser in step 1 and step 2 to determine censorship status using the same decoding techniques proposed by Zhu et al. as described above \shortcite{zhu-etal:2013}.
Altogether, around 41 thousand posts are collected, in which 952 posts (2.28\%) are censored by Weibo.  
In our ongoing work, we are comparing the accuracy of the classifier on posts that are automatically removed vs. those removed by humans. The results will be reported in the future publications.  

\begin{table}[t] 
    \centering
     \resizebox{200pt}{!}{%
    \begin{tabular}{c|c|c}

    \textbf{topic} &  \textbf{censored} & \textbf{uncensored}\\
        \hline
        {cultural revolution} & 55 & 66   \\
         \hline
        {human rights}& 53 & 71 \\
         \hline
        {family planning}& 14 & 28  \\
         \hline
        {censorship \& propaganda}& 32 & 56 \\
         \hline
        {democracy}& 119 & 107 \\
         \hline
        {patriotism}& 70 & 105 \\
         \hline
        {China}& 186 & 194 \\
         \hline
        {Trump}& 320 & 244 \\
         \hline
        {Meng Wanzhou}& 55 & 76 \\
        \hline
        {kindergarten abuse}& 48 & 5  \\
        \hline
        {\textbf{Total}} & \textbf{952} & \textbf{952}

    \end{tabular}
    }
    \caption{Data collected by scraper for classification}
    \label{tab:scraped-table}
\end{table}

We would like to emphasize that while the data collection methods could be seen as recreating a keyword search, the scraping pipeline also deciphers the logic in discovering censorship on Weibo. 

\subsection{Sample Data}

\begin{figure}[H]
\hspace*{-0.75in}
\centering
\includegraphics[width=0.66\textwidth]{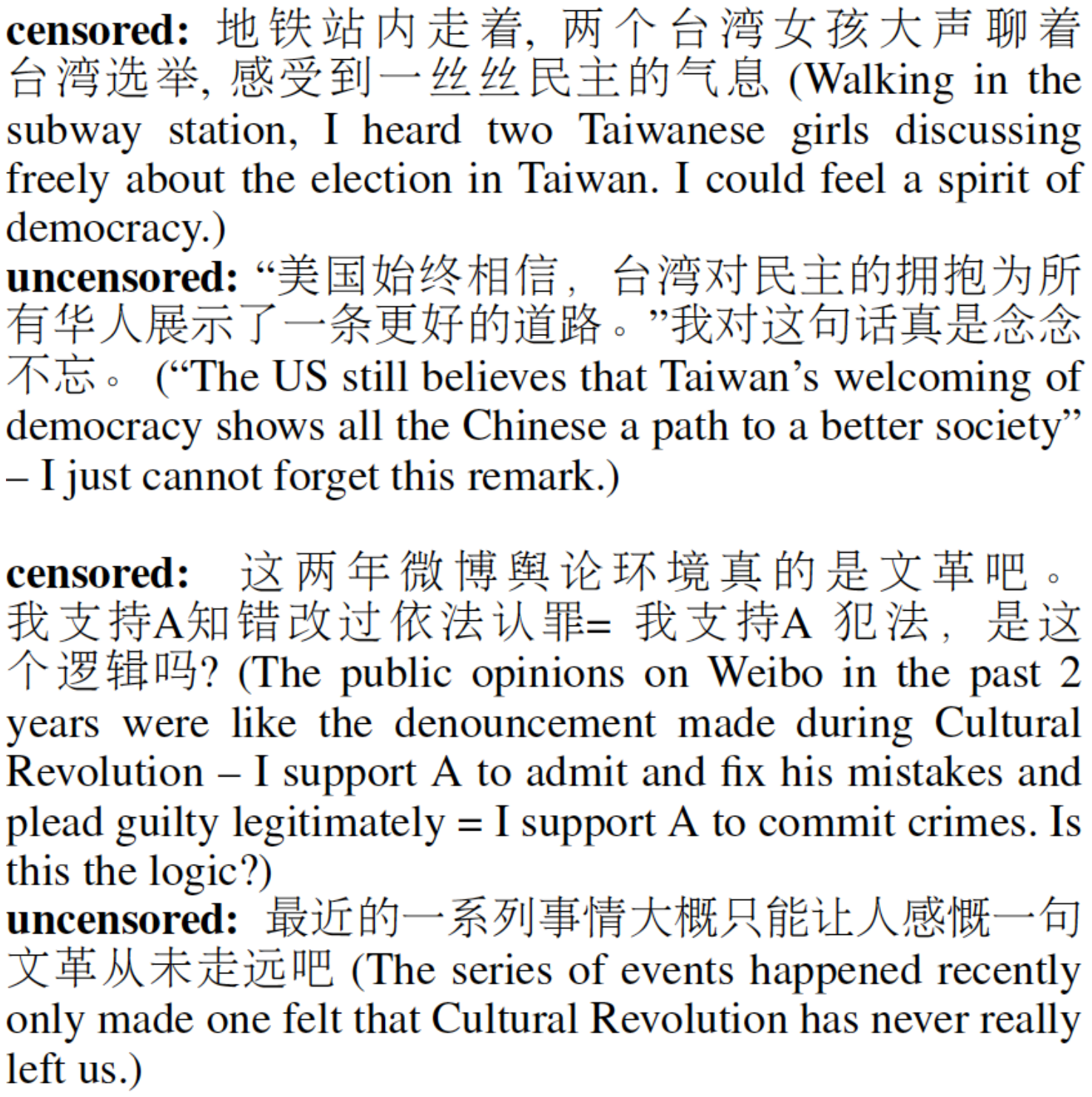} 
\vspace*{-45mm}
\caption{Sample Data}\label{fig:sample}
\end{figure}

Figure \ref{fig:sample} shows several examples selected randomly from our dataset. Each pair of censored and uncensored posts contains the same sensitive keyword.
\vspace{-35mm}

%

\section{Crowdsourcing Experiment} 
\begin{figure}[H]
\hspace*{-0.75in}
\centering
\includegraphics[width=0.66\textwidth]{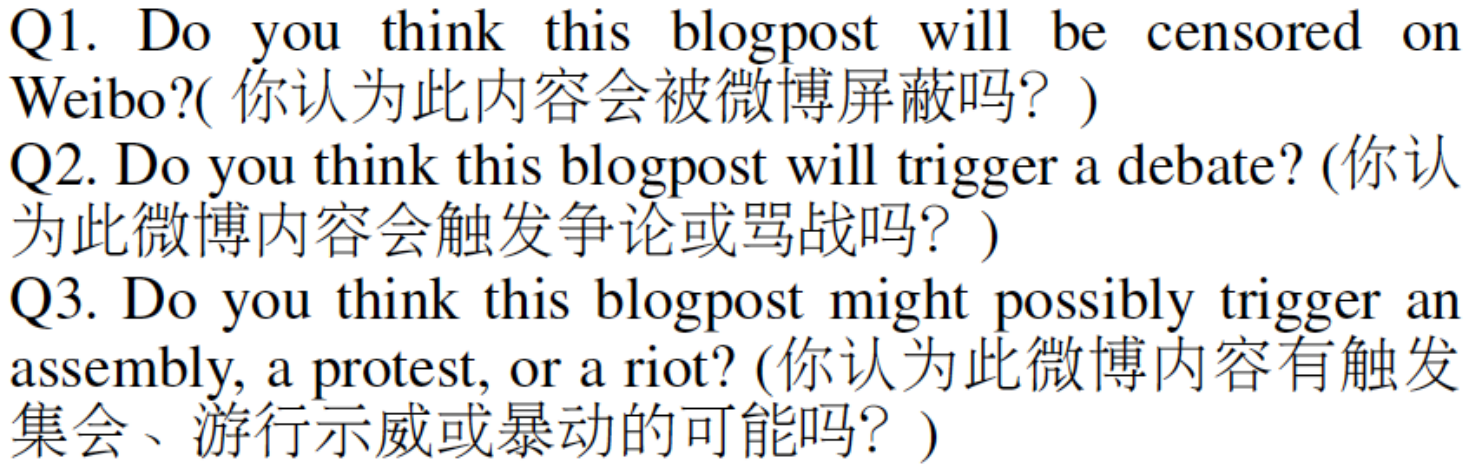}
\vspace*{-65mm}
\caption{Crowdsourcing experiment: Three survey questions.}\label{fig:questions}
\end{figure}

A balanced corpus is created. The uncensored posts of each dataset are randomly sampled to match with the number of their censored counterparts (see Table \ref{tab:scraped-table} and Table \ref{tab:jed-table}).
We select randomly a subset of the data collected by the web scraper to construct surveys for crowdsourcing experiment. 
The surveys ask participants three questions (see Figure \ref{fig:questions}).
\vspace{-60mm}


Sample questions are included in Appendix\footnote{\url{http://msuweb.montclair.edu/~feldmana/publications/aaai20_appendix.pdf}}.
Question 1 explores how humans perform on the task of censorship classification; question 2 explores whether a blogpost is controversial; question 3 serves as a way to explore in our data the concept of Collective Action Potential (CAP) suggested by King et al.\shortcite{king-etal:2013}. According to King et al. \shortcite{king-etal:2013}, Collective Action Potential is the potential to cause collective action such as protest or organized crowd formation outside the Internet. 
Participants can respond either Yes or No to the 3 questions above. A total of 800 blogposts (400 censored and 400 uncensored) are presented to 37 different participants through a crowdsourcing platform Witmart\footnote{\url{http://www.witmart.com}} in 8 batches (100 blogposts per batch). Each blogpost is annotated by 6 to 12 participants. The purpose of this paper is to shed light on the “knowledge gap” between censors and normal weibo users about censorable content. We believe weibo users are aware of potential censorable content but are not “trained” enough to avoid or identify them. The results are summarized in Table \ref{tab:crowdsourcing-table}.

\begin{table}[H]
   \centering
   \resizebox{200pt}{!}
    {%
    \begin{tabular}{l|c|c}
    \textbf{question} & \textbf{censored} & \textbf{uncensored}\\
    \hline
    Q1 (censorship) & \makecell{Yes: 23.83\% \\ No: 76.17\% }& \makecell{Yes: 16.41\% \\ No: 83.59\% }   \\
 \hline
    Q2 (controversy) & \makecell{Yes: 61.02\% \\ No: 38.98\% }& \makecell{Yes: 55.37\% \\ No: 44.63\% }   \\
  \hline
    Q3 (CAP) & \makecell{Yes: 6.13\% \\ No: 93.87\% }& \makecell{Yes: 3.93\% \\ No: 96.07\% }   \\
    \hline
      \end{tabular}
      }
    \caption{\label{tab:crowdsourcing-table} Crowdsourcing results}
\end{table}

The annotation results are intuitive -- participants tend to see censored blogposts as more controversial and more likely to trigger action in real life than the uncensored counterpart. 

We obtain a Fleiss's kappa score for each question to study the inter-rater agreement. Since the number and identity of participants of each batch of the survey are different, we obtain an average Fleiss' kappa from the results of each batch. The Fleiss' kappa of questions 1 to 3 are 0.13, 0.07, and 0.03 respectively, which all fall under the category of slight agreement. 

We hypothesize that since all blogposts contain sensitive keyword(s), the annotators choose to label a fair amount of uncensored blogposts as controversial, and even as likely to be censored or cause action in real life. This might also be the reason of the low agreement scores -- the sensitive keywords might be the cause of divided opinions. 


Regarding the result of censorship prediction, 23.83\% of censored blogposts are correctly annotated as censored, while 83.59\% of 
uncensored blogposts are correctly annotated as uncensored. This result suggests that participants tend to predict a blogpost to survive censorship on Weibo, despite the fact that they can see the presence of controversial element(s) in a blogpost as suggested by the annotation results of question 2.    This suggests that detecting censorable content is a non-trivial task and humans do not have a good intuition (unless specifically trained, perhaps) what material is going to be censored. It might be true that there is some level of subjectivity form human censors. We believe there are commonalities among censored blogposts that pass through the “subjectivity filters” and such commonalities could be the linguistic features that contribute to our experiment results (see sections \ref{sec:features} and \ref{sec:classifications}).

\section{Feature Extraction} \label{sec:features}
To build an automatic classifier, we first extract features from both our scraped data and Zhu et al.'s dataset. While the datasets we use are different from that of Ng et al.~\shortcite{kei-nlp4if:2018} and Ng et al.~\shortcite{ng-etal-2019-neural}, some of the features we extract are similar to theirs. We include CRIE features (see below) and the number of followers feature that are not extracted in Ng et al.~\shortcite{kei-nlp4if:2018}'s work. 

\begin{table}
    \centering
    \resizebox{200pt}{!}{%
    \begin{tabular}{c|c|c}
    \textbf{topic} &  \textbf{censored} & \textbf{uncensored}\\
    \hline
        \makecell{cultural revolution} & 19 & 29   \\
        \hline
        \makecell{human rights}& 16 & 10 \\
        \hline
        \makecell{family planning}& 4 & 4  \\
        \hline
        \makecell{censorship \& propaganda}& 47 & 38 \\
        \hline
        \makecell{democracy}& 94 & 53 \\
        \hline
        \makecell{patriotism}& 46 & 30 \\
        \hline
        \makecell{China}& 300 & 458 \\
        \hline
        \makecell{Bo Xilai}& 8 & 8\\
        \hline
        \makecell{brainwashing}& 57 & 3 \\
        \hline
        \makecell{emigration}& 10 & 11 \\
        \hline
        \makecell{June 4th}& 2 & 5 \\
        \hline
        \makecell{food \& env. safety}& 14 & 17\\
        \hline
        \makecell{wealth inequality}& 2 & 4 \\
        \hline
        \makecell{protest \& revolution}& 4 & 5 \\
        \hline
        \makecell{stability maintenance}& 66 & 28 \\
        \hline
        \makecell{political reform}& 12 & 9 \\
        \hline
        \makecell{territorial dispute}& 73 & 75 \\
        \hline
        \makecell{Dalai Lama}& 2 & 2 \\
        \hline
        \makecell{HK/TW/XJ issues}& 2 & 4 \\
        \hline
        \makecell{political dissidents}& 2 & 2 \\
        \hline
        \makecell{Obama}& 8 & 19 \\
        \hline
        \makecell{USA}& 62 & 59 \\
        \hline
        \makecell{communist party}& 37 & 10 \\
        \hline
        \makecell{freedom}& 12 & 11 \\
        \hline
        \makecell{economic issues}& 31 & 37 \\
        \hline
        \makecell{\textbf{Total}} & \textbf{930} & \textbf{930} \\ \end{tabular}}
    \caption{\label{tab:jed-table}Data extracted from Zhu et al. \shortcite{zhu-etal:2013}'s dataset for classification}
\end{table}

\subsection{Linguistic Features} \label{sec:ling-features} 
We extract 5 sets of linguistic features from both datasets (see below) -- the LIWC features, the CRIE features, the sentiment features, the semantic features, and word embeddings. We are interested in the LIWC and CRIE features because they are purely linguistic, which aligns with the objective of our study. Also, some of the LIWC features extracted from Ng et al. \shortcite{ng2018detecting}'s data have shown to be useful in classifying censored and uncensored tweets.  
 \vspace{-3mm}
\paragraph{LIWC features}
The English Linguistic Inquiry and Word Count (LIWC) \cite{penne-etal:2007,penne-etal:2015} is a program that analyzes text on a word-by-word basis, calculating percentage of words that match each language dimension, e.g., pronouns, function words, social processes, cognitive processes, drives, informal language use etc. Its lexicon consists of approximately 6400 words divided into categories belong to different linguistic dimensions and psychological processes. LIWC  builds on previous research establishing strong links between linguistic patterns and personality/psychological state.
We use a version of LIWC developed for Chinese by Huang et al.~\shortcite{huang-etal:2012} 
 to extract the frequency of word categories. Altogether we extract 95 features from LIWC.  One important feature of the LIWC lexicon is that categories form a tree structure hierarchy. Some features subsume others.
 \vspace{-3mm}
\paragraph{Sentiment features}
We use BaiduAI\footnote{\url{https://ai.baidu.com}} to obtain a set of sentiment scores for each post. BaiduAI's sentiment analyzer is built using deep learning techniques based on data found on Baidu, one of the most popular search engines and encyclopedias in mainland China. It outputs a positive sentiment score and a negative sentiment score which sum to 1.

\vspace{-3mm}
\paragraph{CRIE features}
We use the Chinese Readability Index Explorer (CRIE) \cite{Sung2016}, a text analysis tool developed for measuring the readability of a Chinese text based on the its linguistic components. Its internal dictionaries and lexical information are developed based on dominant corpora such as the Sinica Tree Bank. CRIE outputs 50 linguistic features (see Appendix\footnote{\url{http://msuweb.montclair.edu/~feldmana/publications/aaai20_appendix.pdf}}), such as word, syntax, semantics, and cohesion in each text or produce an aggregated result for a batch of texts. CRIE can train and categorize texts based on their readability levels. We use the textual-features analysis for our data and derive readability scores for each post in our data. These scores are mainly based on descriptive statistics.


\vspace{-3mm}
\paragraph{Semantic features}
We use the Chinese Thesaurus developed by Mei~\shortcite{mei:1984} and extended by HIT-SCIR\footnote{Harbin Institute of Technology Research Center for Social Computing and Information Retrieval.} to extract semantic features. The structure of this  semantic dictionary is similar to WordNet, where words are divided into 12 semantic classes 
and each word can belong to one or more classes. It can be roughly compared to the concept of word senses. We derive a semantic ambiguity feature by dividing the number of words in each post by the number of semantic classes in it. 
\vspace{-3mm}
\paragraph{Frequency \& readability}
We compute the average frequency of characters and words in each post using Da~\shortcite{da:2004}\footnote{\url{http://lingua.mtsu.edu/chinese-computing/statistics/}}'s work and Aihanyu's CNCorpus \footnote{\url{http://www.aihanyu.org/cncorpus/index.aspx}} respectively. For words with a frequency lower than 50 in the reference corpus, we count it as 0.0001\%. 
It is intuitive to think that a text with less semantic variety and more common words and characters is relatively easier to read and understand. We derive a Readability feature by taking the mean of character frequency, word frequency and word count to semantic classes described above. It is assumed that the lower the mean of the 3 components, the less readable a text is. In fact, these 3 components are part of Sung et al. \shortcite{sung-et-al:2015}'s readability metric for native speakers on the word level and semantic level. 
\vspace{-3mm}
\paragraph{Word embeddings}
Word vectors are trained using the word2vec tool \cite{mikolov1,mikolov2} on 300,000 of the latest Chinese articles\footnote{\url{https://dumps.wikimedia.org/zhwiki/latest/}} on Wikipedia. A 200-dimensional vector is computed for each word of each blogpost. The vector average of each blogpost is the sum of word vectors divided by the number of vectors. The 200-dimensional vector average are used as features for classification.

\subsection{Non-linguistic Features}

\paragraph{Followers}
The number of followers of the author of each post is recorded and used as a feature for classification. 

\section{Classification} \label{sec:classifications}
Features extracted from the balanced datasets (See Table 1 and Table 3) are used for classifications. Although the amount of uncensored blogposts significantly outnumber censored in real-life, such unbalanced corpus might be more suitable for anomaly detection. 
All numeric values of the features have been standardized before classification.
We use a multilayer perceptron (MLP) classifier to classify instances into censored and uncensored. A number of classification experiments using different combinations of features are carried out. 

Best performances are achieved using the combination of CRIE, sentiment, semantic, frequency, readability and follower features (i.e. all features but LIWC and word embeddings) (see Table \ref{tab:results-table}).  The feature selection is performed using random sampling. 
As a result 77 features are selected that perform consistently well across the datasets. We call these features the best features set. (see \url{https://msuweb.montclair.edu/~feldmana/publications/aaai20_appendix.pdf} for the full list of features). 

We vary the number of epochs and hidden layers. The rest of the parameters are set to default -- learning rate of 0.3, momentum of 0.2, batch size of 100, validation threshold of 20. Classification experiments are performed on 1) both datasets 2) scraped data only 3) Zhu et al.'s data only. Each experiment is validated with 10-fold cross validation. We report the accuracy of each model in Table \ref{tab:results-table}.
It is worth mentioning that using the LIWC features only, or the CRIE features only, or the word embeddings only, or all features excluding the CRIE features, or all features except the LIWC and CRIE features all result in poor performance of below 60\%. 
Besides MLP, we also use the same sets of features to train classifiers using Naive Bayes, Logistic, and Support Vector Machine. However, the performances are all below 65\%.

\begin{table*}[t]
    \centering
   \resizebox{320pt}{!}{%
    \begin{tabular}{l|c|c|c|c|c|c}
    \textbf{dataset} & \textbf{N} &  \textbf{H} &  \textbf{features} & \textbf{A} & \textbf{P} &\textbf{R}\\
    \hline
        \multicolumn{4}{c}{majority class baseline} & 49.98 \\
        \hline
        \multicolumn{4}{c}{human baseline} & 23.83 \\
        \hline
         \multicolumn{4}{c}{SVM, Naive Bayes, Logistic Regression} & 65\% \\
        \hline
        scraped & 500 & 50,50,50  & BFS  & 80.36 & \makecell{c: 79.3 \\ u: 81.0} & \makecell{c: 75.1 \\ u: 77.4} \\
        \hline
        scraped & 800 & 60,60,60 & BFS & 80.2 & \makecell{c: 81.5 \\ u: 75.5} & \makecell{c: 79.4 \\ u: 79.2}  \\
        \hline
        Zhu et al's & 800 & 50,7 & BFS & 87.63 & \makecell{c: 85.9 \\ u: 86.0} & \makecell{c: 86.0 \\ u: 86.0} \\
        \hline
        Zhu et al's & 800 & 30,30 & BFS & 86.18 & \makecell{c: 85.3 \\ u: 86.2} & \makecell{c: 87.4 \\ u: 86.2} \\
        \hline
        both & 800 & 60,60,60 & BFS & 75.4 & \makecell{c: 72.0 \\ u: 70.6} & \makecell{c: 71.9 \\ u: 75.4}  \\
        \hline
        both & 500 & 50,50,50 & BFS & 73.94 & \makecell{c: 70.7 \\ u: 71.1} & \makecell{c: 73.5 \\ u: 72.0}  \\
        \hline
        scraped & 800 & 30,30,30 & \makecell{all except LIWC \\ \& word embeddings} & 72.95 & \makecell{c: 71.8 \\ u: 73.0} & \makecell{c: 75.6 \\ u: 73.0}  \\
        \hline
        Zhu et al's & 800 & 60,60,60 & \makecell{all except LIWC \\ \& word embeddings} & 70.64 & \makecell{c: 89.2 \\ u: 76.3} & \makecell{c: 45.4 \\ u: 69.9} \\
        \hline
        both & 500 & 40,40,40 & \makecell{all except LIWC \\ \& word embeddings} & 84.67  & \makecell{c: 80.2 \\ u: 79.9} & \makecell{c: 80.4 \\ u: 79.9}  \\
        \hline
        both & 800 & 20,20,20 & \makecell{all except LIWC \\ \& word embeddings} & 88.50  & \makecell{c: 84.6 \\ u: 87.0} & \makecell{c: 80.3 \\ u: 81.7}  \\
        \hline
        both & 800 & 30,30,30 & \makecell{all except LIWC \\ \& word embeddings} & 87.04  & \makecell{c: 86.6 \\ u: 86.0} & \makecell{c: 81.9 \\ u: 85.8}  \\
        \hline
        both & 800 & 50,50,50 & \makecell{all except LIWC \\ \& word embeddings} & 87.24  & \makecell{c: 83.5 \\ u: 84.6} & \makecell{c: 80.8 \\ u: 82.6}  \\
    \end{tabular}}\caption{\label{tab:results-table} MLP classification results. N = number of epochs, H = number of nodes in each hidden layer, A = accuracy, P = precision, R = recall, BFS = best features set, c = censored, u = uncensored}
\end{table*}

\begin{figure*}
 \begin{center}  
 \includegraphics[scale=0.5]{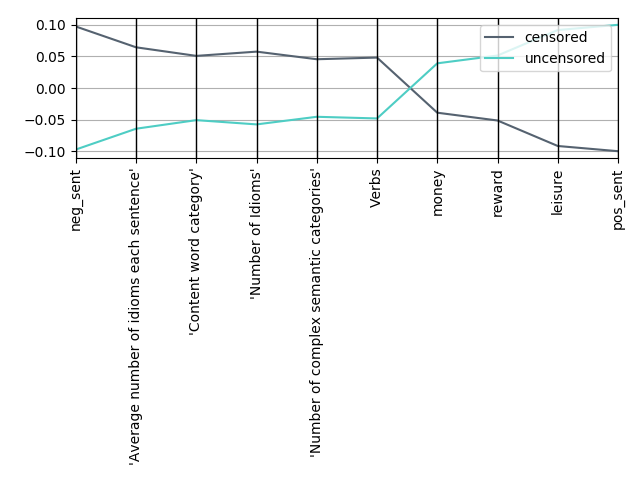}
 \caption{Parallel Coordinate Plots of the top 10 features that have the greatest difference in average values}\label{fig:parallel}
\end{center} 
\end{figure*} 

\section{Discussion and Conclusion} \label{sec:discussion}  
 Our best results are over 30\% higher than the baseline and about 60\% higher than the human baseline obtained through crowdsourcing, which shows that our classifier has a greater censorship predictive ability compared to human judgments. The classification on both datasets together tends to give higher accuracy using at least 3 hidden layers. However, the performance does not improve when adding additional layers (other parameters being the same). Since the two datasets were collected differently and contain different topics, combining them together results in a richer dataset that requires more hidden layers to train a better model. 
It is worth noting that classifying both datasets using the best features set decreases the accuracy, while using all features but LIWC improves the classification performance.  The reason for this behavior could be an existence of consistent differences in the LIWC features between the datasets. Since the LIWC features in the best features set (see Appendix \url{https://msuweb.montclair.edu/~feldmana/publications/aaai20_appendix.pdf}) consist of mostly word categories of different genres of vocabulary (i.e. grammar and style agnostic), it  might suggest that the two datasets use vocabularies differently. Yet, the high performance obtained excluding the LIWC features shows that the key to distinguishing between censored and uncensored posts seems to be the features related to writing style, readability, sentiment, and semantic complexity of a text.  

\begin{figure}[H]
\hspace*{-2.0in}
\centering
\includegraphics[width=1.1\textwidth]{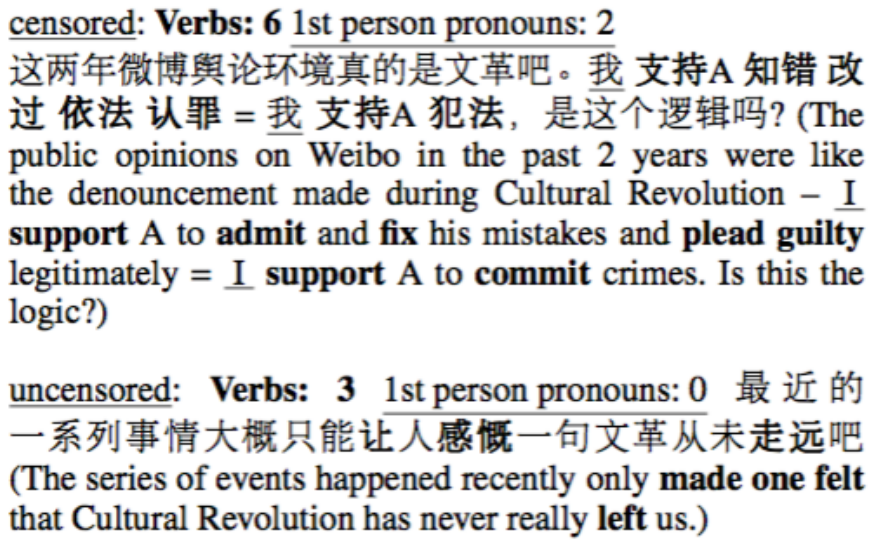} 
\vspace*{-105mm}
\caption{Examples of blogposts annotated by CRIE.}\label{fig:crie}
\end{figure}

 Figure \ref{fig:crie} shows two blogposts annotated by CRIE with number of verbs and number of first person pronoun features. 
\vspace{-105mm}


 
To narrow down on what might be the best features that contribute to distinguishing censored and uncensored posts, we compare the mean of each feature of the two classes (see Figure \ref{fig:parallel}). The 6 features distinguish censored from uncensored are: \\

\indent 1. negative sentiment \\
\indent 2. average number of idioms in each sentence \\
\indent 3. number of content word categories \\
\indent 4. number of idioms \\
\indent 5. number of complex semantic categories \\
\indent 6. verbs \\

On the other hand, the 4 features that distinguish uncensored from censored are: \\

\indent 1. positive sentiment \\
\indent 2. words related to leisure \\ 
\indent 3. words related to reward \\
\indent 4. words related to money \\ 

This might suggest that the censored posts generally convey more negative sentiment and are more idiomatic and semantically complex in terms of word usage. According to King et al. \shortcite{king-etal:2013}, Collective Action Potential, which is related to a blogpost's potential of causing riot or assembly in real-life, is the key determinant of a blogpost getting censored. Although there is not a definitive measure of this construct, it is intuitive to relate a higher average use of verbs to a post that calls for action. 


On the other hand, the uncensored posts might be in general more positive in nature (positive sentiment) and include more content that talks about neutral matters (money, leisure, reward).

We further explore how the use of verbs might possibly affect censorship by studying the types of verbs used in censored and uncensored blogposts. We extracted verbs from all blogposts by using the Jieba Part-of-speech tagger \footnote{\url{https://github.com/fxsjy/jieba}}. We then used the Chinese Thesaurus described in Section \ref{sec:ling-features} to categorize the verbs into 5 classes: Actions, Psychology, Human activities, States and phenomena, and Relations. However, no significant differences have been found across censored and uncensored blogposts. A further analysis on verbs in terms of their relationship with actions and arousal can be a part of future work. 

Since the focus of this paper is to study the linguistic content of blogposts, rather than rate of censorship, we did not employ technical methods to differentiate blogposts that have different survival rates. Future work could be done to investigate any differences between blogposts that get censored at different rates. In our ongoing work, we are comparing the accuracy of the classifier on posts that are automatically removed vs. those removed by humans. The results will be reported in the future publications.  

To conclude,  our work shows that there are linguistic fingerprints of censorship, and it is possible to use linguistic properties of a social media post to automatically predict if it is going to be censored. It will be interesting to explore if the same linguistic features can be used to predict censorship on other social media platforms and in other languages.  


\section*{Acknowledgments}
This work is supported by the National Science Foundation under Grant No.: 1704113, Division of Computer and Networked Systems, Secure and Trustworthy Cyberspace (SaTC). We also thank Jed Crandall for sharing  Zhu et al. \shortcite{zhu-etal:2013}'s dataset with us. \\

\bibliography{censorship}

\begin{thebibliography}{}

\bibitem[\protect\citeauthoryear{Bamman, O'Connor, and
  Smith}{2012}]{bamman-etal:2012}
Bamman, D.; O'Connor, B.; and Smith, N.~A.
\newblock 2012.
\newblock {Censorship and deletion practices in Chinese social media}.
\newblock {\em First Monday} 17(3).

\bibitem[\protect\citeauthoryear{Da}{2004}]{da:2004}
Da, J.
\newblock 2004.
\newblock A corpus-based study of character and bigram frequencies in chinese
  e-texts and its implications for chinese language instruction.
\newblock In Zhang, P.; Tianwei, X.; and Xu, J., eds., {\em The studies on the
  theory and methodology of the digitalized Chinese teaching to foreigners:
  Proceedings of the Fourth International Conference on New Technologies in
  Teaching and Learning Chinese.},  501--511.
\newblock Beijing: Tsinghua University Press.

\bibitem[\protect\citeauthoryear{Hiruncharoenvate, Lin, and
  Gilbert}{2015}]{hirun-etal:2015}
Hiruncharoenvate, C.; Lin, Z.; and Gilbert, E.
\newblock 2015.
\newblock {Algorithmically Bypassing Censorship on Sina Weibo with
  Nondeterministic Homophone Substitutions}.
\newblock In {\em Ninth International AAAI Conference on Web and Social Media}.

\bibitem[\protect\citeauthoryear{Huang \bgroup et al\mbox.\egroup
  }{2012}]{huang-etal:2012}
Huang, C.-L.; Chung, C.; Hui, N.~K.; Lin, Y.-C.; Seih, Y.-T.; Lam, B.~C.; Chen,
  W.-C.; Bond, M.; and Pennebaker, J.~H.
\newblock 2012.
\newblock The development of the chinese linguistic inquiry and word count
  dictionary.
\newblock {\em Chinese Journal of Psychology} 54(2):185--201.

\bibitem[\protect\citeauthoryear{ji{\=a}~j{\=u} M{\'e}i}{1984}]{mei:1984}
ji{\=a}~j{\=u} M{\'e}i.
\newblock 1984.
\newblock {\em The Chinese Thesaurus}.

\bibitem[\protect\citeauthoryear{Katti, Katabi, and
  Puchala}{2005}]{katti-etal:2005}
Katti, S.; Katabi, D.; and Puchala, K.
\newblock 2005.
\newblock {Slicing the onion: Anonymous routing without pki.}
\newblock Technical report, MIT CSAIL Technical Report 1000.

\bibitem[\protect\citeauthoryear{King, Pan, and Roberts}{2013}]{king-etal:2013}
King, G.; Pan, J.; and Roberts, M.~E.
\newblock 2013.
\newblock {How Censorship in China Allows Government Criticism but Silences
  Collective Expression}.
\newblock {\em American Political Science Review} 107(2):1--18.

\bibitem[\protect\citeauthoryear{Knockel \bgroup et al\mbox.\egroup
  }{2015}]{knockel-etal:2015}
Knockel, J.; Crete-Nishihata, M.; Ng, J.; Senft, A.; and Crandall, J.
\newblock 2015.
\newblock Every rose has its thorn: Censorship and surveillance on social video
  platforms in china.
\newblock In {\em Proceedings of the 5th USENIX Workshop on Free and Open
  Communications on the Internet.}

\bibitem[\protect\citeauthoryear{Leberknight, Chiang, and
  Wong}{2012}]{leberknight-etal:2012a}
Leberknight, C.~S.; Chiang, M.; and Wong, F. M.~F.
\newblock 2012.
\newblock A taxonomy of censors and anti-censors: Part i-impacts of internet
  censorship.
\newblock {\em International Journal of E-Politics (IJEP)} 3(2).

\bibitem[\protect\citeauthoryear{Lee}{2016}]{lee:2016}
Lee, S.
\newblock 2016.
\newblock Surviving online censorship in china: Three satirical tactics and
  their impact.
\newblock {\em China Quarterly}.

\bibitem[\protect\citeauthoryear{Levin \bgroup et al\mbox.\egroup
  }{2015}]{levin-etal:2015}
Levin, D.; Lee, Y.; L.Valenta; amd V.~Lai, Z.~L.; Lumezanu, C.; Spring, N.; and
  Bhattacharjee, B.
\newblock 2015.
\newblock Alibi routing.
\newblock In {\em Proceedings of the 2015 ACM Conference on Special Interest
  Group on Data Communication.}

\bibitem[\protect\citeauthoryear{McPherson, Shokri, and
  Shmatikov}{2016}]{mcpherson-etal:2016}
McPherson, R.; Shokri, R.; and Shmatikov, V.
\newblock 2016.
\newblock Defeating image obfuscation with deep learning.
\newblock arXiv preprint arXiv:1609.00408.

\bibitem[\protect\citeauthoryear{Mikolov \bgroup et al\mbox.\egroup
  }{2013a}]{mikolov1}
Mikolov, T.; Chen, K.; Corrado, G.; and Dean, J.
\newblock 2013a.
\newblock Efficient estimation of word representations in vector space.
\newblock In {\em Proceedings of Workshop at ICLR}.

\bibitem[\protect\citeauthoryear{Mikolov \bgroup et al\mbox.\egroup
  }{2013b}]{mikolov2}
Mikolov, T.; Sutskever, I.; Chen, K.; Corrado, G.; and Dean, J.
\newblock 2013b.
\newblock Distributed representations of words and phrases and their
  compositionality.
\newblock In {\em Proceedings of NIPS}.

\bibitem[\protect\citeauthoryear{Ng \bgroup et al\mbox.\egroup
  }{2018}]{kei-nlp4if:2018}
Ng, K.~Y.; Feldman, A.; Peng, J.; and Leberknight, C.
\newblock 2018.
\newblock {Linguistic Characteristics of Censorable Language on SinaWeibo}.
\newblock In {\em Proceedings of the 1st Workshop on NLP for Internet Freedom
  held in conjunction with COLING 2018}.

\bibitem[\protect\citeauthoryear{Ng \bgroup et al\mbox.\egroup
  }{2019}]{ng-etal-2019-neural}
Ng, K.~Y.; Feldman, A.; Peng, J.; and Leberknight, C.
\newblock 2019.
\newblock Neural network prediction of censorable language.
\newblock In {\em Proceedings of the Third Workshop on Natural Language
  Processing and Computational Social Science},  40--46.
\newblock Minneapolis, Minnesota: Association for Computational Linguistics.

\bibitem[\protect\citeauthoryear{Ng, Feldman, and
  Leberknight}{2018}]{ng2018detecting}
Ng, K.~Y.; Feldman, A.; and Leberknight, C.
\newblock 2018.
\newblock Detecting censorable content on sina weibo: A pilot study.
\newblock In {\em Proceedings of the 10th Hellenic Conference on Artificial
  Intelligence}.
\newblock ACM.

\bibitem[\protect\citeauthoryear{Pennebaker \bgroup et al\mbox.\egroup
  }{2015}]{penne-etal:2015}
Pennebaker, J.~W.; Boyd, R.~L.; Jordan, K.; and Blackburn, K.
\newblock 2015.
\newblock The development and psychometric the development of psychometric
  properties of liwc.
\newblock Technical report, University of Texas at Austin.

\bibitem[\protect\citeauthoryear{Pennebaker, Booth, and
  Francis}{2017}]{penne-etal:2007}
Pennebaker, J.~W.; Booth, R.; and Francis, M.
\newblock 2017.
\newblock {\em Linguistic Inquiry and Word Count (LIWC2007)}.

\bibitem[\protect\citeauthoryear{Safaka \bgroup et al\mbox.\egroup
  }{2016}]{safaka-etal:2016}
Safaka, I.; Fragouli, C.; ; and Argyraki, K.
\newblock 2016.
\newblock Matryoshka: Hiding secret communication in plain sight.
\newblock In {\em 6th USENIX Workshop on Free and Open Communications on the
  Internet (FOCI 16). USENIX Association.}

\bibitem[\protect\citeauthoryear{Sung \bgroup et al\mbox.\egroup
  }{2015}]{sung-et-al:2015}
Sung, Y.; Chang, T.; Lin, W.; Hsieh, K.; and Chang, K.
\newblock 2015.
\newblock Crie: An automated analyzer for chinese texts.
\newblock {\em Behavior Research Method}.

\bibitem[\protect\citeauthoryear{Sung \bgroup et al\mbox.\egroup
  }{2016}]{Sung2016}
Sung, Y.-T.; Chang, T.-H.; Lin, W.-C.; Hsieh, K.-S.; and Chang, K.-E.
\newblock 2016.
\newblock Crie: An automated analyzer for chinese texts.
\newblock {\em Behavior Research Methods} 48(4):1238--1251.

\bibitem[\protect\citeauthoryear{Tager, Bass, and Lopez}{2018}]{pen-2018}
Tager, J.; Bass, K.~G.; and Lopez, S.
\newblock 2018.
\newblock Forbidden feeds: Government controls on social media in china.
\newblock Technical report, Pen America.

\bibitem[\protect\citeauthoryear{Weinberg \bgroup et al\mbox.\egroup
  }{2012}]{weinberg-etal:2012}
Weinberg, Z.; Wang, J.; Yegneswaran, V.; Briesemeister, L.; Cheung, S.; Wang,
  F.; and Boneh, D.
\newblock 2012.
\newblock Stegotorus: A camouflage proxy for the tor anonymity system.
\newblock {\em Proceedings of the 19th ACM conference on Computer and
  Communications Security}.

\bibitem[\protect\citeauthoryear{Zhu \bgroup et al\mbox.\egroup
  }{2013}]{zhu-etal:2013}
Zhu, T.; Phipps, D.; Pridgen, A.; Crandall, J.~R.; and Wallach, D.~S.
\newblock 2013.
\newblock The velocity of censorship: High-fidelity detection of microblog post
  deletions.
\newblock In {\em Proceedings of the 22Nd USENIX Conference on Security},
  SEC'13,  227--240.
\newblock Berkeley, CA, USA: USENIX Association.

\end{thebibliography}
\bibliographystyle{aaai}

\end{document}